\documentclass{article}

\usepackage[preprint,               
            nonatbib]{neurips_2025}

\usepackage{listings}
\lstdefinestyle{pythonstyle}{
    language=Python,
    basicstyle=\ttfamily\small,
    keywordstyle=\color{blue},
    stringstyle=\color{orange},
    commentstyle=\color{gray},
    numbers=left,
    numberstyle=\tiny,
    numbersep=5pt,
    frame=single,
    framexleftmargin= 1.0em, 
    xleftmargin= 1.0pt,        
    backgroundcolor=\color{white},
    showspaces=false,
    showstringspaces=false,
    showtabs=false,
    tabsize=4,
    captionpos=b,
    breaklines=true,
    breakatwhitespace=false,
                    numbersep=5pt,
                xleftmargin=.25in,
                xrightmargin=.25in
}

\usepackage{tikz}
\usepackage{listofitems}
\tikzstyle{mynode}=[thick,draw=black,circle,minimum size=11]

\usetikzlibrary{arrows.meta} 
\usepackage[outline]{contour} 
\contourlength{1.4pt}

\usepackage{xcolor}
\colorlet{myred}{red!80!black}
\colorlet{myblue}{blue!80!black}
\colorlet{mygreen}{green!60!black}
\colorlet{myorange}{orange!70!red!60!black}
\colorlet{mydarkred}{red!30!black}
\colorlet{mydarkblue}{blue!40!black}
\colorlet{mydarkgreen}{green!30!black}

\tikzset{
  >=latex, 
  node/.style={thick,circle,draw=myblue,minimum size=11,inner sep=0.0,outer sep=0.},
  node in/.style={node,green!20!black,draw=mygreen!30!black,fill=mygreen!25},
  node hidden/.style={node,blue!20!black,draw=myblue!30!black,fill=myblue!20},
  node convol/.style={node,orange!20!black,draw=myorange!30!black,fill=myorange!20},
  node out/.style={node,red!20!black,draw=myred!30!black,fill=myred!20},
  connect/.style={thick,mydarkblue}, 
  connect arrow/.style={-{Latex[length=4,width=3.5]},thick,mydarkblue,shorten <=0.5,shorten >=1},
  node 1/.style={node in}, 
  node 2/.style={node hidden},
  node 3/.style={node out}
}
\def\nstyle{int(\lay<\Nnodlen?min(2,\lay):3)} 

\usepackage[utf8]{inputenc} 
\usepackage[T1]{fontenc}    
\usepackage{hyperref}       
\usepackage{url}            
\usepackage{booktabs}       
\usepackage{amsfonts}       
\usepackage{nicefrac}       
\usepackage{microtype}      
\usepackage{xcolor}         
\usepackage{float}
\usepackage{bbm}

\usepackage{amsmath}
\usepackage{amssymb}
\usepackage{amsthm}
\usepackage{mathrsfs}
\usepackage{subcaption}
\usepackage{caption}

\usepackage{graphicx}
\usepackage{float}
\usepackage[title]{appendix}
\newcommand{\scalefactor}{0.65}

\title{\textbf{Emergence of Structure in Ensembles of Random Neural Networks}}

\author{%
Luca Muscarnera$^{1}$ 
\\
\textbf{ 
\quad Luigi Loreti$^2$ 
\quad Giovanni Todeschini$^3$
}
\\
\textbf{ 
\quad Alessio Fumagalli$^{1}$ 
\quad Francesco Regazzoni$^{1}$
} 
\\ 
\\
$^1$MOX Laboratory, Department of Mathematics, Politecnico di Milano (Italy)
\\
\texttt{ \{luca.muscarnera,alessio.fumagalli,francesco.regazzoni\}@polimi.it }
\\
\\
$^2$ Optopc SA, Chiasso (Switzerland)
\\
\texttt{luigi.loreti@optopc.com}
\\
\\
$^3$ Rebel Dynamics Srl, Cesana Brianza (Italy) 
\\
\texttt{
giovanni.todeschini@rebeldynamics.it
}
}

\begin{document}

\maketitle

\begin{abstract}
   Randomness is ubiquitous in many applications across data science and machine learning. Remarkably, systems composed of random components often display emergent global behaviors that appear deterministic, manifesting a transition from microscopic disorder to macroscopic organization. In this work, we introduce a theoretical model for studying the emergence of collective behaviors in ensembles of random classifiers. We argue that, if the ensemble is weighted 
    through the Gibbs measure defined by adopting the classification loss as an energy, then there exists a finite temperature parameter for the distribution such that the classification is optimal, with respect to the loss (or the energy). Interestingly, for the case in which samples are generated by a Gaussian distribution and labels are constructed by employing a teacher perceptron, we analytically prove and numerically confirm that such optimal temperature does not depend neither on the teacher classifier (which is, by construction of the learning problem, unknown), nor on the number of random classifiers, highlighting the universal nature of the observed behavior. Experiments on the \texttt{MNIST} dataset underline the relevance of this phenomenon in high-quality, noiseless, datasets. Finally, a physical analogy allows us to shed light on the self-organizing nature of the studied phenomenon.
\end{abstract}

\section{Introduction}
The study of Random Neural Networks (also known as Untrained Neural Networks or, in a slightly different context, as Neural Networks with Random Weights \cite{cao2018review}) is inherently entangled to the foundations of the Learning problem \cite{cao2018review,bishop2006pattern}. A neural network can be, in fact, studied as the evolution of a disordered system --- since at initialization the realization of the parameters is a form of "frozen" (or more technically, quenched) disorder, making the network, in fact, random \cite{couillet2022random} --- towards a state in the space of parameters --- reached through any possible training algorithm \cite{amari1993backpropagation} --- where the randomness is diluted in favor of the emergence of some form of structure in the biases and in the weight matrices, which are learned through the observation of data \cite{engel2001statistical}.
In fact, when a Neural Network is initialized, weight matrices are obtained by sampling from random matrices whose entries --- in the classical initialization schemes \cite{kumar2017weight,narkhede2022review} --- lack of a covariance structure which therefore leads to random (agnostic) classification; on the other
 hand, the high accuracies obtained after training suggest the existence of some form of highly structured information encoded into the weights \cite{le1991eigenvalues}.
The emergence  of such information in the parameters of the network allows the performing of many different tasks, often with characteristics that are superior in quality with respect to their human based counterpart \cite{vaswani2017attention,jumper2021highly}.
However, it is still not clear what characterizes the geometry and the distribution of weights in a well-trained neural network \cite{jin2020does,martin2020heavy}. {What  property is responsible for making a certain realization of the parameters more efficient than the others? Is it possible to impose the same form of structure \textit{before} the training, or even as \textit{alternative} training strategy?} These questions, inevitably linked with the complex nature of neural networks \cite{balcazar2002computational, mezard2024spin}, drove us towards the construction of a minimal model of a Random Neural Network where randomness in the system is preserved, and the classification outcome is constructed through a weighted majority voting performed by the randomly initialized neurons of the network. 
The underlying motif of our work can be summarized by the following question: {can a system of many \cite{anderson1972more} \textit{randomly} and independently initialized neurons perform non trivial tasks if the magnitude of their influence on the final classification is informed by the loss  function?} More formally, we propose a framework where the output of a certain (large) number $n$ of randomly initialized perceptrons \cite{block1962perceptron} is combined by  means of a weighted sum wherein the weight $\alpha_{\mathbf w}$ associated to each perceptron with parameter $\mathbf w \in \mathbb R^d$ satisfies the following proportionality relationship
\begin{equation}
\label{eq:rule}
    \alpha_{\mathbf w}(\beta) \propto \exp ( - \beta \mathcal L(\phi (  \mathbf w)))
\end{equation}

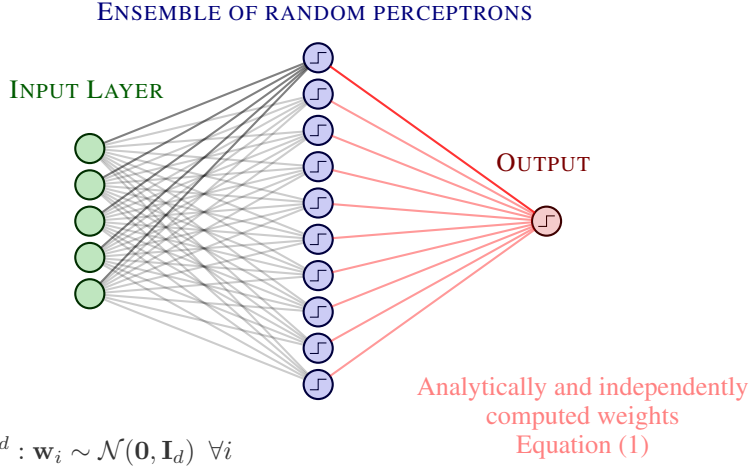
\begin{figure}
    \centering
\begin{tikzpicture}[x=2.2cm,y=1.4cm,scale = 0.23]
  \message{^^JNeural network with arrows}
  \readlist\Nnod{5,10,1} 
  
  \message{^^J  Layer}
  \foreachitem \N \in \Nnod{ 
    \edef\lay{\Ncnt} 
    \message{\lay,}
    \pgfmathsetmacro\prev{int(\Ncnt-1)} 
    \foreach \i [evaluate={\y=1.5*(\N/2-\i); \x=6 * \lay; \n=\nstyle;}] in {1,...,\N}{ 
      
      \node[node \n] (N\lay-\i) at (\x,\y) {
        \ifnum \Ncnt > 1
                \tikz \draw plot coordinates {(0.05,0) (0.1,0.) (0.1,0.1) (0.15,0.1)};            
        \fi
      };
      
      \ifnum\lay>1 
        \foreach \j in {1,...,\Nnod[\prev]}{ 
            \ifnum\Ncnt=2 
                \ifnum \i = 1 
                  \draw[thick, opacity=0.5] (N\prev-\j) -- (N\Ncnt-\i); 
                \fi
                \ifnum \i > 1 
                  \draw[thick, opacity=0.2] (N\prev-\j) -- (N\Ncnt-\i); 
                \fi
             \fi
                \ifnum\Ncnt>2
                  \ifnum \j = 1
                    \draw[thick, opacity=0.8, red] (N\prev-\j) -- (N\Ncnt-\i); 
                  \fi 
                  \ifnum \j > 1
                    \draw[thick, opacity=0.4, red] (N\prev-\j) -- (N\Ncnt-\i); 
                  \fi
            \fi
        }
      \fi 
      
    }
    
  }
  
  \node[above=-120,left = -60,align=center,black, opacity = 0.8] at( N1-1.90 ) {
  $\mathbf W  \in \mathbb R^{n \times d} : \mathbf w_i \sim \mathcal N(\mathbf 0, \mathbf I_d) \ \ \forall i$
  };
  \node[above=-80,left = -80,align=center,red, opacity = 0.5] at( N3-1.90 ) {
  Analytically and independently
  \\ computed weights
  \\
  Equation (\ref{eq:rule})
  };

  \node[above=10,align=center,mygreen!60!black] at (N1-1.90) {
  \textsc{Input Layer}
  };
  \node[above=5,align=center,myblue!60!black] at (N2-1.90) {
  \textsc{
  Ensemble of random perceptrons
  }
  };
  \node[above=10,align=center,myred!60!black] at (N\Nnodlen-1.90) {
  \textsc{
  Output
  }
  };
  
\end{tikzpicture}
    \caption{
        Model of the proposed architecture, with null biases. The gray weights are sampled from a normal distribution, while red weights are computed analytically by the rule described in Equation \ref{eq:rule}.
        The highlighted subnetwork represents the flow of information from the input, to one randomly constructed classifier and to the output.
        The hidden layer can be considered as an ensemble of random elementary neural networks (perceptrons) whose output is then combined through the analytically computed weights, which do not require a training phase to be estimated, with each weight in the last layer being independent from the others.
    Both hidden and output neurons adopt the $\operatorname{sign}$ activation function.
    }
    \label{fig:pippo}
\end{figure}

with $\mathcal L$ being the loss  function associated to each perceptron (mapping each classifier into an empirical risk measure computed  on  observed data), $\beta$ being a parameter that allows to modulate the influence of low loss configurations, denoted  from now on as the inverse temperature 
 of the model (being, thus, a positive value) and $\phi$ the \textit{realization} map that associates a weight vector $\mathbf w$ to  the perceptron $\mathbf x \mapsto \operatorname{sign}(\mathbf w^T \mathbf x)$. This choice of the weight is consistent with the idea of defining a 
 Gibbs measure  \cite{huang2008statistical} over the possible classifiers, with $\mathcal L$ playing the role of an energy.
 The ensemble is taking the form
\begin{equation*}
    \hat y_{ens}(\mathbf W, \mathbf x) := \operatorname{sign}\left(
    \sum_i^n \alpha_{\mathbf{w}_i} (\beta)\operatorname{sign}(\mathbf w_i^T \mathbf x)
    \right)
    = \operatorname{sign}\left( 
        \boldsymbol{\alpha}_{\mathbf W}^T(\beta) \operatorname{sign}\left( 
            \mathbf W \mathbf x
        \right)
    \right),
\end{equation*}
where $\mathbf W \in \mathbb R^{n \times d}$ is the realization of a suitably defined random matrix, with statistically independent entries and with the $i-$th row  denoted as $\mathbf w_i$.
From a probabilistic perspective, we can therefore think that the vector of the parameters of the network, defined concatenating $\operatorname{flat}(\mathbf W)$ and $\boldsymbol{\alpha}_{\mathbf W}(\beta)$, may be imagined as a non isotropic random vector since the vector of weights on the last layer is statistically non independent (even in a deterministic relationship, if one assumes that the data that induces the loss is fixed) from  the entries of the random matrix $\mathbf W$ that models the action of the first layer. This  article aims to prove, empirically and analytically when possible, that this specific structure in the statistics of weights allows the definition of a training free learning paradigm, { 
which may be --- after further development --- computationally efficient especially 
 in the context of dedicated hardware able to process massively parallel workflows, due to the avoidance of the seriality bottleneck which is intrinsic in more classical training methods }.

Our paper is structured in the following way: in Section \ref{sec:related} we briefly discuss the literature on Random Neural Networks, with a sharp focus on the methods that involve stochasticity as a tool to perform inference, in Section \ref{sec:anres} we discuss the main analytical result based on the hypothesis of Gaussian uncorrelated data and the corresponding experimental validation setup, proving in both cases that the $\beta$ parameter is independent of the specific properties of the 
 task that we aim to learn and in Section \ref{sec:mnist} we describe an experiment on organic data --- the \texttt{MNIST} \cite{lecun2010mnist}  dataset ---  underlining a non obvious compatibility with the proposed analytical model. Finally, in Section \ref{sec:phys} a physical analogy sheds a light on the underlying dynamics of the classification and the implications of our results are critically  discussed. 
\section{Related work}\label{sec:related}
To the best of the authors' knowledge, the approach of imposing a covariance structure to the  weights of a Neural Network has not been  extensively tackled in literature. The survey \cite{narkhede2022review} presents some examples of {data-driven} initializations, where some information from data is distilled into the weights in order to achieve better convergence properties or higher accuracy. Nevertheless, literature appears to be lacking of examples where the absence of independence among the neural networks' weights is actively exploited to perform inference, without a further training phase. Some similarities could be found in the conceptual ideas behind Reservoir Computing \cite{yan2024emerging}, where  properties of random weight matrices  are exploited to perform useful tasks, or Extreme Learning Machines \cite{wang2022review}, where random  weights allow to perform a sort of uninformed feature selection. In 
 an extremely broad sense, architectural bias constitutes a form of statistical dependence among weights; for instance, a Convolutional Neural Network may be interpreted as a Multi Layer Perceptron where, in each layer, some weights are constrained to be equal \cite{fukushima1969visual} forcing therefore a non-isotropic correlation structure among weights (which are of course not part of the actual parameters of the network, since they simply would be redundant).  Nonetheless, our research in literature \cite{saxe2011random,cox2011beyond,giryes2016deep,gallicchio2020deep} suggests that this may be the first work to actively discuss the injection of statistical dependence between random weights of the untrained neural network to perform  inference.
 
\section{Training-free Gaussian data classification}\label{sec:anres}
Consider the case of Gaussian (isotropic) data where we imagine that  labels are generated by  means of an unknown teacher \cite{hinton2015distilling} vector $\mathbf w_* \in \mathbb R^d$ (a fixed $d-$dimensional vector) which induces the definition of two half-spaces, one where  data is classified as $+1$ and one where data  is classified as $-1$ according to the sign of the dot product of a query vector and $\mathbf w_*$.
Let $\mathbf x \sim \mathcal N(\mathbf 0,\mathbf I_d)$ be an  isotropic multivariate normal random variable. Suppose to observe a dataset $\{ (\mathbf x_i, y_i) \}_{i = 1}^N \subset \mathbb R^d \times \{-1,1\}$ where each sample $\mathbf x_i$ is a realization of $\mathbf x$ and $y_i = y_*(\mathbf x_i)  \ \forall i$ where $ y_*(\mathbf x) =  \operatorname{sign}(\mathbf w_*^T \mathbf x)$. Our objective, in this context, is to study wether for a given family of random classifiers of the form
\begin{equation*}
    \hat y_i(\mathbf x) = \operatorname{sign}(\mathbf w_i^T \mathbf x)
    \hspace{2em} i =1,2,...,n
\end{equation*}
there exists a (untrained) weighted majority voting scheme to combine their outputs such that the classification performed by the ensemble significantly outperforms the best (according to a certain loss function $\mathcal L$) classifier in the family. More formally, we study the case where $\mathbf w_1, \dots, \mathbf w_n$ are the rows of a random matrix of gaussian entries $\mathbf W$ (where $W_{ij} \sim \mathcal N(0,1) \ \forall i \in [n],j \in [d]$) and thus, due to the radial symmetry of the isotropic multivariate normal distribution, each random classifier defines a random separating hyperplane passing through the origin by sampling uniformly over the possible normal vectors. As we mentioned, the weights of the majority voting are defined according to a Gibbs measure over the different quenched random functionals, leading to the ensemble classification
\begin{equation*}
    \hat y = \operatorname{sign}\left( 
        \frac{1}{
        Z_n(\beta)
        }
        \sum_i^n \exp
        ( - \beta \mathcal L(\hat y_i))
        \hat y_i
    \right),
\end{equation*}
where the loss function $\mathcal L$ plays the role of the energy in the Gibbs measure,  the $\operatorname{sign}$ function is intended --- with a slight abuse of notation --- as the operator that maps a function $f : \mathbb R^d \rightarrow \mathbb R$ into its projection on the codomain $\{-1,1\}$ with $(\operatorname{sign}(f))(\mathbf x) = \operatorname{sign}(f(\mathbf x)) \ \forall \mathbf x \in \mathbb R^d$ and the Partition Function $Z_n(\beta) = \sum_i^n \exp( - \beta \mathcal L(\hat y_i))$ acts as a normalization constant which, being composed with a $\operatorname{sign}$, serves the only purpose of making the limit $\beta \rightarrow \infty$ well defined. It is important to underline that the possibility of omitting $Z_n(\beta)$ for finite $\beta$ makes the weights of the last layer independent on each other.
Being $\beta$ the only degree of freedom of the Gibbs measure, the problem reduces to the study of the existence of an optimal parameter $\beta^*$ (and thus, an optimal inverse temperature of the system) for which, we conjecture, the ensemble manifests an optimal prediction quality (with respect to the loss function $\mathcal L$), possibly unpredictable from the observation of single individuals from the population of classifiers, exhibiting therefore an emerging behavior. To motivate (and  formalize) this idea, we propose --- firstly --- a conceptual experiment; consider, in fact, the case where the number of random classifiers is infinite, and thus we obtain
\begin{equation*}
    \begin{split}
     \lim_{n \rightarrow +\infty} 
    \operatorname{sign} \left(
    \frac{1}{Z_n(\beta)}
    \sum_i^n \exp
        ( - \beta \mathcal L(\hat y_i))
        \hat y_i
    \right)
    \! \!
    = \operatorname{sign} \left(
    \frac{1}{Z_\infty(\beta)}
    \int_{\mathbb R^d} 
    \!  \! \!
    \mu( d \mathbf w)
    \exp
        ( - \beta \mathcal L(\phi(\mathbf w_i)))
        \phi(\mathbf w_i)
    \right)
    \end{split}
\end{equation*}
with $\mu$ being the probability measure over the weights $\mathbf w$, which we defined as normally distributed and $Z_\infty (\beta) =   \int_{\mathbb R^n}  \mu( d \mathbf w)
    \exp
        ( - \beta \mathcal L(\phi(\mathbf w_i)))
$. Hence, being $\operatorname{supp}(\mu) = \mathbb R^d$ we obtain the following equality based on the limit of the softmin \cite{gao2017properties} function
%
\begin{equation*}
    \begin{split}
    &  \lim_{\beta \rightarrow +\infty}
    \lim_{n \rightarrow +\infty} 
    \operatorname{sign}\left(
    \frac{1}{Z_n(\beta)}
    \sum_i^n \exp
        ( - \beta \mathcal L(\hat y_i))
        \hat y_i
        \right) = 
    \phi \left(
    \mathop{\text{argmin}}_{\mathbf w \in \operatorname{supp}(\mu)} \mathcal  L(\phi(\mathbf w))
    \right)
     = \phi(\mathbf w_*)
     = y_*
    \end{split}
\end{equation*}
meaning that, in the infinite classifiers scenario $\beta^* = +\infty$, since $\mathbf w_* \in \operatorname{supp}(\mu)$ and the Gibbs measure converges to a Dirac Delta centered in the global minimizer.  Nonetheless, in a context when $n$ is instead finite and thus we observe only a sample from the population of classifiers the probability measure $\mu$ is approximated by $\hat \mu$ given by
\begin{equation*}
    \hat \mu(\mathbf w) = n^{-1} \sum_i^n \delta(\mathbf w - \mathbf w_i)
\end{equation*}
which, unlike $\mu$, has the zero measure support $\operatorname{supp}(\hat \mu) = \bigcup_{i=1}^n \{ \mathbf w_i \}$ with therefore $0$ probability of containing $\mathbf w_*$  and with
\begin{equation*}
    \begin{split}
    \lim_{\beta \rightarrow +\infty}
    \operatorname{sign} \left(
    \frac{1}{Z_n(\beta)}
    \sum_i^n \exp
        ( - \beta \mathcal L(\hat y_i))
        \hat y_i
    \right)
    &
    =
    \phi \left( \mathop{\text{argmin}}_{\mathbf w \in \{\mathbf w_i\}_{i=1}^n } 
    \mathcal L(\phi(\mathbf w_i))
    \right).
    \end{split}
\end{equation*}
The implication is direct; if in the infinite classifiers scenario the $\beta \rightarrow \infty$ limit retrieves the global minimizer, in the finite case the minimizer of the sample is obtained. On the other hand, if we consider the curve
\begin{equation*}
    \gamma(\beta) = \operatorname{sign} \left(
    \frac{1}{Z_n(\beta)}
    \sum_i^n \exp
        ( - \beta \mathcal L(\hat y_i))
        \hat y_i
    \right)
    \hspace{2em} \text{with } \beta \in [0,\infty)
\end{equation*}
we note that it is defined as a $1-$dimensional curve in the functional space $\{ f : \mathbb R^d \rightarrow \{-1,1\}\}$,
and that its trajectory lies in the image of the convex  hull of the set of the sampled classifiers through the $\operatorname{sign}$ operator, and it intersects the set of sampled classifiers for $\beta \rightarrow \infty$ in its minimizer as we explained before. Due to this intersection, we obtain the following upper bound
\begin{equation*}
    \min_{\beta} \mathcal L(\gamma(\beta)) \le \min_{i} \mathcal L(\phi(\mathbf w_i)) = \lim_{\beta \rightarrow \infty } \mathcal L(\gamma(\beta)),
\end{equation*}
suggesting that, unless $\mathcal L(\gamma(\beta))$ is monotonically decreasing with respect to $\beta$, it should exists a minimizer $\beta^*$ for which the ensemble classification outperforms the best classifier in the ensemble.
The intuition behind our analysis relies in the idea that if $\beta$ acts modulating the amount of information in each classifier that enters the ensemble classification (with the already discussed trivial behavior for $\beta \rightarrow \infty$ and the agnostic average for $\beta \approx 0$), the ensemble could still benefit ---  in terms of accuracy ---  from employing information contained in every classifier, rather than neglecting suboptimal ones (where optimality is defined adopting the sample as domain).

Following a semi-empirical approach, a first numerical experiment is conducted  to evaluate if the model exhibits a non-trivial minimum in its loss for a finite $\beta$. The experiment is prepared by considering a random dataset of $N$ samples defined as
\begin{equation*}
    D = \{ (\mathbf x_i, y_i)_i \}_{i = 1}^N \hspace{2em} \text{with } \mathbf x_i \sim \mathcal N(\mathbf 0, \mathbf I_d) \text{ and } y_i = \operatorname{sign}(\mathbf w_*^T \mathbf x),
\end{equation*}
with $\mathbf w_*$ a randomly selected teacher vector, and then constructing the ensemble of random perceptrons by sampling the Gaussian random matrix $\mathbf W \in \mathbb R^{n \times d}$. To understand the challenging nature of the problem of performing correct classification through randomization, we perform, prior to the experiment, a preliminary study of the distribution of the loss function among the different classifiers.

\begin{figure}
    \centering
    \includegraphics[scale = \scalefactor]{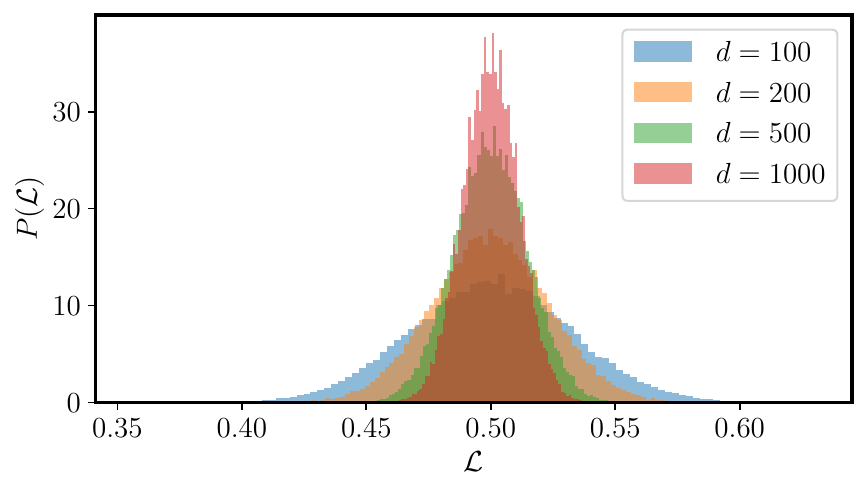}
    \caption{Distribution of the composition of the loss function and the random variable $\mathbf w \sim \mathcal N(\mathbf 0, \mathbf I_d)$, as the dimensions increase it becomes exponentially harder to sample classifier that significantly outperform the average case}
    \label{fig:conc}
\end{figure}

The result of the simulation can be observed in Figure \ref{fig:conc}, where the distribution of the images of random parameters of the perceptron through the loss function is studied. The observed concentration phenomenon (with the empirical estimate of the variance vanishing for large $d$) suggests that in a infinite-dimensional scenario every random classifier is agnostic with respect to the loss function,  concentrating steeply around the average case. Connecting to our discussion on the behavior of the ensemble for $\beta \rightarrow +\infty$, the implication of this property is that in a high dimensional but finite classifiers case we expect that the ensemble behaves equally for $\beta = 0$ and $\beta = +\infty$, with both cases performing a classification equivalent --- in quality --- to a random guess. To investigate the behavior of the system for finite values of $\beta$, we analyzed the ensemble loss for different values of $\beta$, in order to study the image (hereafter referred as \textit{loss profile}) of the curve $\gamma$ through the loss function.

%
%
\begin{figure}
    \centering
    \includegraphics[scale = \scalefactor]{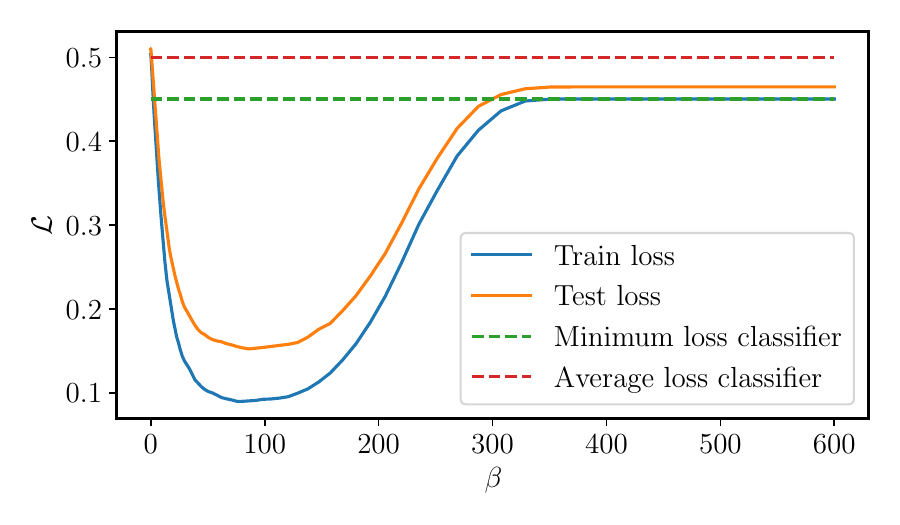}
    \caption{Loss of the ensemble for different values of $\beta$. The profile of the function shows a clear structure with a distinct minimmum, which significantly overcomes the concentration of  the loss in high dimensional data. }
    \label{fig:gaussian}
\end{figure}

The result of the experiment can be observed in Figure \ref{fig:gaussian}. We note some interesting facts on the behavior of the ensemble; first, one may notice that the loss profile  appears to be "smooth", with a clearly continuous structure, with vanishing fluctuations, a clear minimum at $\beta \approx 100$ and thus with a trend that is perfectly consistent with our expectations. As a second, remarkable, point the  ensemble shows generalization abilities; the network appears to be able to generalize also on unseen data, despite of the high dimensional geometry  of Isotropic Gaussian data (which, due to the full-rankness of its covariance, has its support on the whole $\mathbb R^d$), with the position of  the minimum coinciding for both test and training data (note that in this context the term  "training set" is an abuse of notation, since data are not actively used in a real training phase but only employed in the direct construction of the last layer). A third, interesting aspect of our simulation, is underlined by the emergence of quality in the classification despite the high error of the best classifier in the ensemble ($\mathcal L(\mathbf w_{best}) \approx 0.45$). This last  property, is in our opinion, at the  core of our work; { how could we justify this emerging characteristics in an ensemble of  \textit{completely random classifiers}?} {Under \textit{which conditions} such emerging behavior manifests?}
To uncover the  nature of the phenomenon a further analysis was conducted on $\beta^*$ under the assumption that a characterization of the optimal inverse temperature could reveal the intrinsic mechanics of the model. Thus, a second experiment was conducted by numerically estimating $\beta^*$ for different parameters of the model (varying, therefore, the number of classifiers $n$, the dimensionality of data $d$ and the number $N$ of sample that are observed).

%
%

\begin{figure}
    \centering
    \begin{subfigure}[b]{0.25\textwidth}
        \includegraphics[scale =\scalefactor]{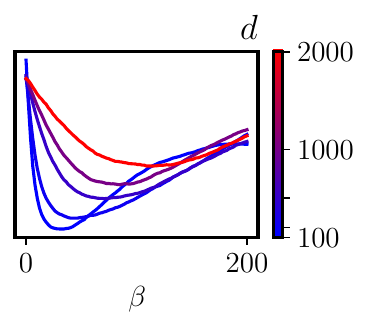}
        \caption{Loss profile for different dimensionalities}
        \label{fig:sub1}
    \end{subfigure}
    \hspace{2.0em}
    \begin{subfigure}[b]{0.25\textwidth}
        \includegraphics[scale =\scalefactor]{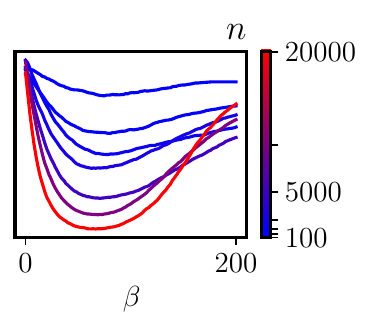}
        \caption{Loss profile for different number of classifiers}
        \label{fig:sub2}
    \end{subfigure}
    \hspace{2.0em}
    \begin{subfigure}[b]{0.25\textwidth}
        \includegraphics[scale =\scalefactor]{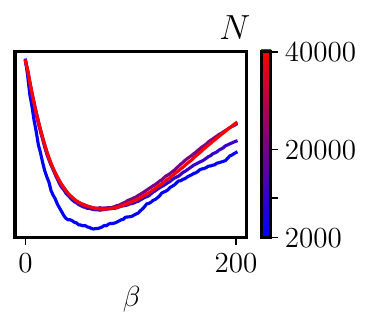}
        \caption{Loss Profile for  different  sizes of the "Training" set}
        \label{fig:sub3}
    \end{subfigure}
    \caption{Structure of the loss profiles in function of perturbations of $d$,  $n$ and $N$. The color scale indicates an increase of the fixed value from blue to red. The position of the optimal inverse temperature appears to be influenced solely by $d$}
    \label{fig:three_subplots}
\end{figure}

Figure \ref{fig:three_subplots} shows the result of the experiment. Each trial is performed by varying a specific quantity (such as $d$) while keeping the other two fixed (such as $n,N$). The empirical evidence suggests that the position of  the minimum is not influenced  neither by the number of classifiers $n$ nor by the number of datapoints $N$, depending solely on the number of dimensions  of the problem $d$. On the other hand, the number of classifiers shows a clear impact on the quality of the classification, highlighting that what we denoted  as "emergence  of structure" is naturally more relevant for large ensembles of  classifiers. In light of these considerations, a final numerical experiment was constructed to study wether the specific choice of the teacher vector $\mathbf w_*$ could have some effect on the position and the quality of the minimum. We generate a single dataset (with a given seed) of datapoints $\mathbf X$ and a quenched set of classifier described by the random matrix $\mathbf W$ and then generate different teacher vectors, one for each different seed, following a multivariate normal distribution (which we remember, lead to teachers that are uniform in direction). The resulting loss profile is obtained, for each tuple $(\mathbf X, \mathbf W, \mathbf w_*^{(i)})$.

%
%
\begin{figure}
    \centering
    \includegraphics[scale = \scalefactor]{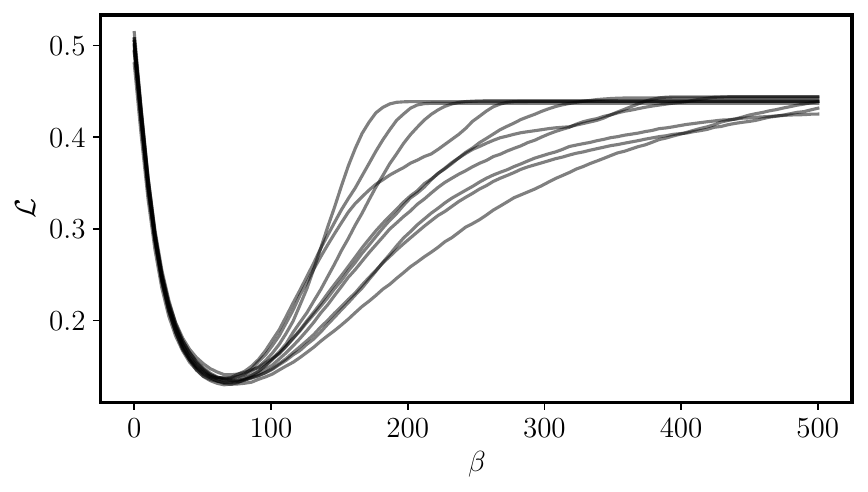}
    \caption{Profile of the loss for different teachers classifiers. Despite of the fluctuations, the position of the minimum is virtually constant among the different simulations }
    \label{fig:teachers}
\end{figure}

In Figure \ref{fig:teachers} we see the result of the simulation for $10$ different teacher vectors. What we can conclude, from our experiments, is that while different factors may impact the shape of the loss profile, the position of the minimum (which is the inverse temperature $\beta$ at which optimal "learning" happens) 
appears to be depending only on the number of dimensions $d$. Moreover, in Figure \ref{fig:teachers} we observe that the quality of the solution obtained by reaching the minimum appears to be invariant with respect to the structure of the teacher classifier, suggesting that the phenomenon is disentangled from the nature of the specific task that has been learned by the model.


The experimental result can be further validated by the construction of an asymptotic  model of the loss profile, that may be obtained by means of the Central Limit Theorem, which explains the loss of the model for a given value of $\beta$.

A demonstration (see Appendix \ref{sec:dimostrazione} for further details on the construction of this result) shows in fact that  the position of the optimal $\beta$ for classification is  approximated by the maximizer of the real function
\begin{equation}
    \label{eq:math_model}
    \xi(\beta) =  
\frac{
    \exp \left\{ \frac{1}{2}\left( 
              \frac{\beta}{\pi \sqrt{d - 2}} 
            \right)^2
    \right \}
     \beta 
    \left( 
              \frac{1}{\pi \sqrt{d - 2}} 
    \right)^2
    }{
    \sqrt{ 
    \exp 
    \left(
              \frac{ 2 \beta^2  }{\pi^2 (d - 2)} 
    \right) 
    -
    4
         \exp \left\{ \left( 
              \frac{\beta}{\pi \sqrt{d - 2}} 
            \right)^2
    \right \}
    \beta^2
    \left( 
              \frac{1}{\pi \sqrt{d - 2}} 
    \right)^4
    }
    },
\end{equation}
which depends solely on $d$, confirming the empirical evidence of independence (up to fluctuations) of the optimal inverse temperature for learning from the structure of the learning task.
%
%
\begin{figure}
    \centering
    \includegraphics[scale = \scalefactor]{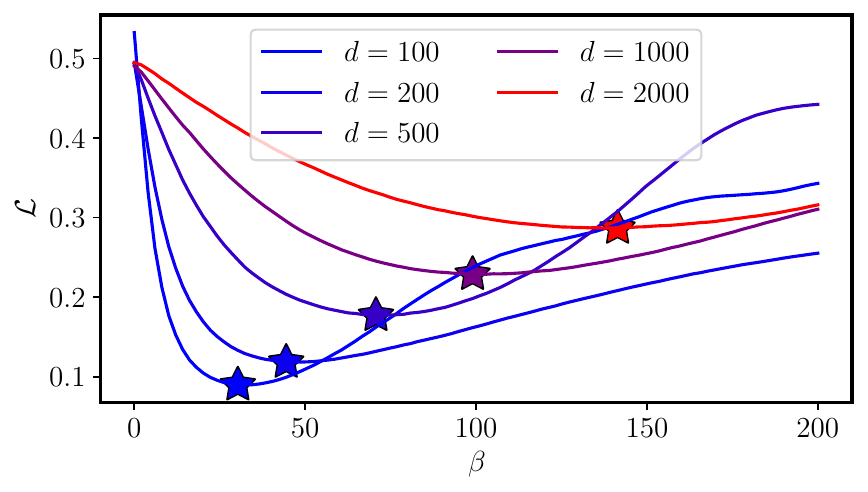}
    \caption{Predictions ($\star$) of $\beta^*$ using the numerical maximization of the function presented in \ref{eq:math_model}, for different values of $d$. The model appears to be able to predict the position of the minimum on the loss curve}
    \label{fig:optimalbeta}
\end{figure}

A further simulation was performed to confirm our analysis, as shown in Figure \ref{fig:optimalbeta}. For each considered dimensionality, the maximizer of $\xi$ (computed through a grid search algorithm) and the empirical loss profile are compared. Despite the approximate nature of the proposed model, the reconstruction appears to be consistent with the empirical observation. Interestingly, a detailed computation (described briefly in Appendix \ref{sec:num_an}) shows that the maximization of $\xi$ admits the closed form
\begin{equation*}
    \beta^* = \pi \sqrt{d - 2},
\end{equation*}
which provides a solution for the optimal learning inverse temperature in the Gaussian case.

\section{Numerical experiments on the \texttt{MNIST} dataset}

Naturally, the behavior of our model on Gaussian data represents a very specific scenario in terms of data geometry. The adoption of an isotropic model of data forces, in fact, symmetries that may be non optimal or even unnatural in describing the structure of real world data. In light of this consideration, it is interesting  to investigate wether the proposed model is able or not to provide a realistic description of the behavior of the Random Neural Network with organic data, which is often, by contrast, highly structured \cite{sclocchi2025phase, poggio2017and}.

Hence, a further numerical investigation was performed on the \texttt{MNIST} dataset. Such dataset offers an interesting playground for testing our hypothesis; the data is, in fact, highly structured, providing well defined clusters \cite{hein2005intrinsic} and a low intrinsic dimension 
 \cite{spigler2020asymptotic}. In our experiment we constructed $N  =  100$ random classifiers, using as a target vector the parity of the digit ($y_i = +1$ if $\mathbf x_i$ represents an even digit, otherwise $y = -1$ if the represented digit is odd). We employed as loss function the $0\text{-}1$ Loss, as we did in the Gaussian model, by estimating the sample average of misclassified samples over the whole training set, composed of $60 000$ digits.

%
%
\begin{figure}[H]
    \centering
    \includegraphics[scale=\scalefactor]{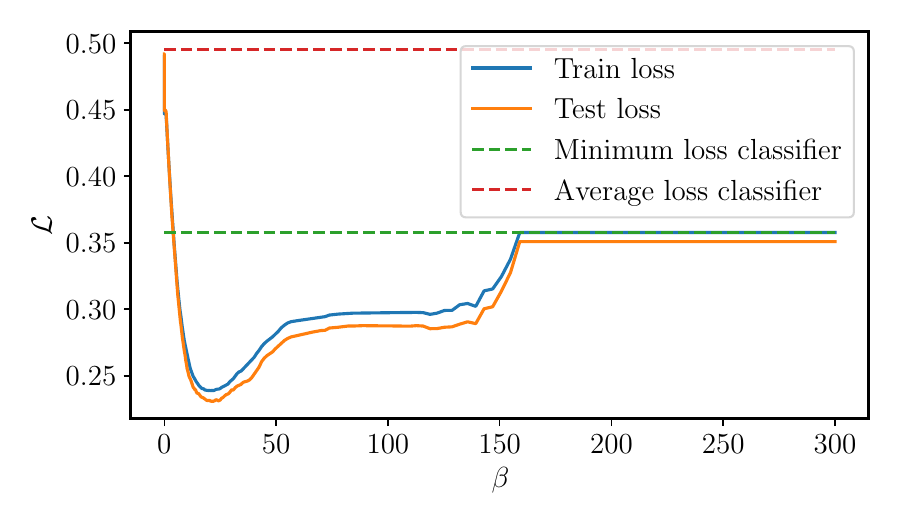}
    \caption{Different values of $\beta$ are considered in this experiment with the \texttt{MNIST} dataset. The profile of the loss has a clear minimum in $\beta \approx 20$,  which is able to  outperform the best classifier}
    \label{fig:mnistplot}
\end{figure}

The result of the simulation can be observed in Figure \ref{fig:mnistplot}. 
Interestingly, we note a similar structure of the curve with respect to the Gaussian model, with a clear minimum for $\beta \approx 20$.

Furthermore, we note that the ensemble generalizes, as may be noted from the loss profile for training and test loss.  We interpret this result as a consequence of the low dimensional structure of the \texttt{MNIST} dataset, which also explains the relatively low best classifier loss despite of the high dimensionality of the dataset ($d = 784$).

\begin{figure}
    \centering
    \includegraphics[scale  =  \scalefactor]{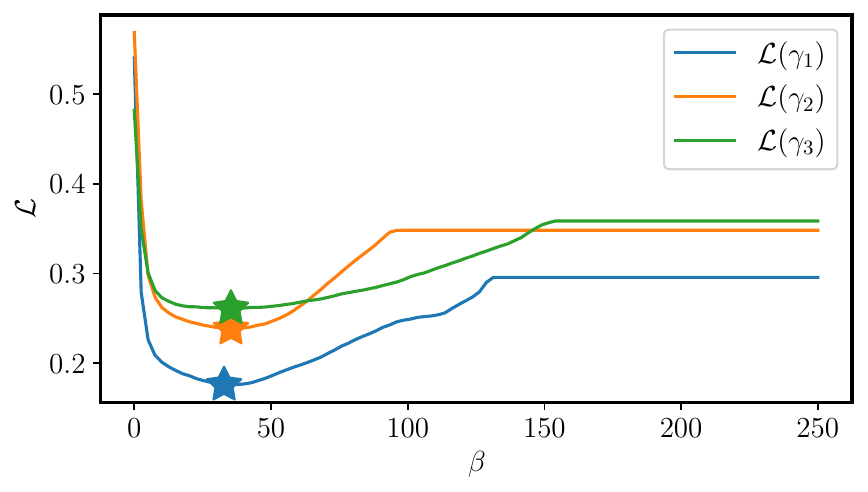}
    \caption{Three different loss profiles based on the \texttt{MNIST} dataset. Stars ($\star$) represent the numerically computed optimal inverse temperature $\beta^*$}
    \label{fig:mnist_three}
\end{figure}

The experiment suggests that the property that we observed for Gaussian data may be in some sense universal, showing  similar features in distributions that differ from the simpler isotropic scenario. However, we still have to verify that the value of the optimal inverse temperature $\beta^*$ does not, approximately, depend on the nature of the training task. To validate this idea we constructed three different problems, with
%
%
\begin{align*}
    y_1(\mathbf x) & = +1 \text{ if the digit is even, } -1 \text{ otherwise}\\  
    y_2(\mathbf x) & = +1 \text{ if the digit is less or equal than } 5 \text{, } -1 \text{ otherwise} \\  
    y_3(\mathbf x) & = \operatorname{sign}(\mathbf w_*^T \mathbf x) \hspace{2em} \mathbf w_* \sim \mathcal N(\mathbf 0, \mathbf I).
\end{align*}
The loss profiles are then plotted for the three problems, as shown in Figure \ref{fig:mnist_three}.  Despite of the different nature of the tasks the optimal inverse temperature $\beta^*$ coincide. This represents an important feature of our model, highlighting the connection of the Isotropic case with more general classes of data distributions.

%
%
\label{sec:mnist}
\section{Physical interpretation, future work and conclusion}
\label{sec:phys}

It is worth noting that the proposed ensemble admits a meaningful physical interpretation. Suppose in fact that an infinitely large ensemble of non-interacting perceptrons $\{P_i\}_{i = 1}^{\infty}$ is employed in the classification. The state $\mathcal S(P)$ of each perceptron is represented by the realization of its weights, that can be one of the rows of the quenched random matrix $\mathbf W$. If the system is "heated" at temperature $1/\beta$ we could imagine that the number of classifiers whose state is $\mathbf w$ is given by the following proportionality relationship
\begin{equation*}
    \# \left\{ P : \mathcal S(P) = \mathbf w \right\}
    \propto \exp( - \beta \mathcal L(\mathbf w))
\end{equation*}
with the result becoming more precise as the size of the ensemble of classifiers grows. If in our model one interprets neurons not as individual perceptrons, but rather as representatives of perceptrons in the same state (given by the random realization of the weights) belonging to the larger ensemble then the analytically computed weights represent --- in the infinite size limit --- the cardinality of the equivalence class associated to each perceptron. This viewpoint on the model highlight how the observed phenomenon could be theoretically treated as an example of self-organization, with the further constraint of non-interaction among agents, a part from the initial consensus  on the choice of the random matrix $\mathbf W$.

{ 
As a final consideration, the theoretical possibility of accelerating the inference has been investigated, showing that under mild assumptions it is possible to merge the predictions by averaging classifiers in the parameter space rather than in the functional space, enabling the construction of a single perceptron classifier that approximates the family of random perceptrons (See Appendix \ref{sec:inference}).  A further analysis of this idea could be fruitful in the context of constructing algorithms based on our work.
}

The natural evolution of this work is to analyze the case in which the perceptron is replaced by more sophisticated examples of random neural networks, such as random multilayer perceptrons or even random convolutional layers, to understand if it is possible to observe this  emerging mechanism also in richer models. Moreover, we shall analyze also if it could be possible to observe similar results with noisy observation  of the loss function (for instance, when the sample mean is computed on batches of the dataset) to move our theoretical analysis into a more practical algorithm. We also argue that this example could be significative in stimulating novel techniques of pretraining for larger  models based on the injection of a covariance structure into the random  parameters of the network. Finally, we want to highlight that the computational cost associated to the analytical construction of weights may be very high for traditional parallel computing systems.  

{ 
We believe that the  methods, which was now proposed under a theoretical lens, could become an effective training method in the context of a dedicated hardware architecture able to exploit the embarrassingly parallel nature of our proposal.
}
\newpage

\bibliography{refs}

\begin{thebibliography}{10}

\bibitem{cao2018review}
W.~Cao, X.~Wang, Z.~Ming, and J.~Gao, ``A review on neural networks with random weights,'' {\em Neurocomputing}, vol.~275, pp.~278--287, 2018.

\bibitem{bishop2006pattern}
C.~M. Bishop and N.~M. Nasrabadi, {\em Pattern recognition and machine learning}, vol.~4.
\newblock Springer, 2006.

\bibitem{couillet2022random}
R.~Couillet and Z.~Liao, {\em Random matrix methods for machine learning}.
\newblock Cambridge University Press, 2022.

\bibitem{amari1993backpropagation}
S.-i. Amari, ``Backpropagation and stochastic gradient descent method,'' {\em Neurocomputing}, vol.~5, no.~4-5, pp.~185--196, 1993.

\bibitem{engel2001statistical}
A.~Engel, {\em Statistical mechanics of learning}.
\newblock Cambridge University Press, 2001.

\bibitem{kumar2017weight}
S.~K. Kumar, ``On weight initialization in deep neural networks,'' {\em arXiv preprint arXiv:1704.08863}, 2017.

\bibitem{narkhede2022review}
M.~V. Narkhede, P.~P. Bartakke, and M.~S. Sutaone, ``A review on weight initialization strategies for neural networks,'' {\em Artificial intelligence review}, vol.~55, no.~1, pp.~291--322, 2022.

\bibitem{le1991eigenvalues}
Y.~Le~Cun, I.~Kanter, and S.~A. Solla, ``Eigenvalues of covariance matrices: Application to neural-network learning,'' {\em Physical review letters}, vol.~66, no.~18, p.~2396, 1991.

\bibitem{vaswani2017attention}
A.~Vaswani, N.~Shazeer, N.~Parmar, J.~Uszkoreit, L.~Jones, A.~N. Gomez, {\L}.~Kaiser, and I.~Polosukhin, ``Attention is all you need,'' {\em Advances in neural information processing systems}, vol.~30, 2017.

\bibitem{jumper2021highly}
J.~Jumper, R.~Evans, A.~Pritzel, T.~Green, M.~Figurnov, O.~Ronneberger, K.~Tunyasuvunakool, R.~Bates, A.~{\v{Z}}{\'\i}dek, A.~Potapenko, {\em et~al.}, ``Highly accurate protein structure prediction with alphafold,'' {\em nature}, vol.~596, no.~7873, pp.~583--589, 2021.

\bibitem{jin2020does}
G.~Jin, X.~Yi, L.~Zhang, L.~Zhang, S.~Schewe, and X.~Huang, ``How does weight correlation affect generalisation ability of deep neural networks?,'' {\em Advances in Neural Information Processing Systems}, vol.~33, pp.~21346--21356, 2020.

\bibitem{martin2020heavy}
C.~H. Martin and M.~W. Mahoney, ``Heavy-tailed universality predicts trends in test accuracies for very large pre-trained deep neural networks,'' in {\em Proceedings of the 2020 SIAM International Conference on Data Mining}, pp.~505--513, SIAM, 2020.

\bibitem{balcazar2002computational}
J.~L. Balc{\'a}zar, R.~Gavalda, and H.~T. Siegelmann, ``Computational power of neural networks: A characterization in terms of kolmogorov complexity,'' {\em IEEE Transactions on Information Theory}, vol.~43, no.~4, pp.~1175--1183, 2002.

\bibitem{mezard2024spin}
M.~M{\'e}zard, ``Spin glass theory and its new challenge: structured disorder,'' {\em Indian Journal of Physics}, vol.~98, no.~11, pp.~3757--3768, 2024.

\bibitem{anderson1972more}
P.~W. Anderson, ``More is different: Broken symmetry and the nature of the hierarchical structure of science.,'' {\em Science}, vol.~177, no.~4047, pp.~393--396, 1972.

\bibitem{block1962perceptron}
H.-D. Block, ``The perceptron: A model for brain functioning. i,'' {\em Reviews of Modern Physics}, vol.~34, no.~1, p.~123, 1962.

\bibitem{huang2008statistical}
K.~Huang, {\em Statistical mechanics}.
\newblock John Wiley \& Sons, 2008.

\bibitem{lecun2010mnist}
Y.~LeCun, C.~Cortes, and C.~Burges, ``Mnist handwritten digit database,'' {\em ATT Labs [Online]. Available: http://yann.lecun.com/exdb/mnist}, vol.~2, 2010.

\bibitem{yan2024emerging}
M.~Yan, C.~Huang, P.~Bienstman, P.~Tino, W.~Lin, and J.~Sun, ``Emerging opportunities and challenges for the future of reservoir computing,'' {\em Nature Communications}, vol.~15, no.~1, p.~2056, 2024.

\bibitem{wang2022review}
J.~Wang, S.~Lu, S.-H. Wang, and Y.-D. Zhang, ``A review on extreme learning machine,'' {\em Multimedia Tools and Applications}, vol.~81, no.~29, pp.~41611--41660, 2022.

\bibitem{fukushima1969visual}
K.~Fukushima, ``Visual feature extraction by a multilayered network of analog threshold elements,'' {\em IEEE Transactions on Systems Science and Cybernetics}, vol.~5, no.~4, pp.~322--333, 1969.

\bibitem{saxe2011random}
A.~M. Saxe, P.~W. Koh, Z.~Chen, M.~Bhand, B.~Suresh, and A.~Y. Ng, ``On random weights and unsupervised feature learning.,'' in {\em Icml}, vol.~2, p.~6, 2011.

\bibitem{cox2011beyond}
D.~Cox and N.~Pinto, ``Beyond simple features: A large-scale feature search approach to unconstrained face recognition,'' in {\em 2011 IEEE International Conference on Automatic Face \& Gesture Recognition (FG)}, pp.~8 -- 15, 04 2011.

\bibitem{giryes2016deep}
R.~Giryes, G.~Sapiro, and A.~M. Bronstein, ``Deep neural networks with random gaussian weights: A universal classification strategy?,'' {\em IEEE Transactions on Signal Processing}, vol.~64, no.~13, pp.~3444--3457, 2016.

\bibitem{gallicchio2020deep}
C.~Gallicchio and S.~Scardapane, ``Deep randomized neural networks,'' in {\em Recent Trends in Learning From Data: Tutorials from the INNS Big Data and Deep Learning Conference (INNSBDDL2019)}, pp.~43--68, Springer, 2020.

\bibitem{hinton2015distilling}
G.~Hinton, O.~Vinyals, and J.~Dean, ``Distilling the knowledge in a neural network,'' {\em arXiv preprint arXiv:1503.02531}, 2015.

\bibitem{gao2017properties}
B.~Gao and L.~Pavel, ``On the properties of the softmax function with application in game theory and reinforcement learning,'' {\em arXiv preprint arXiv:1704.00805}, 2017.

\bibitem{sclocchi2025phase}
A.~Sclocchi, A.~Favero, and M.~Wyart, ``A phase transition in diffusion models reveals the hierarchical nature of data,'' {\em Proceedings of the National Academy of Sciences}, vol.~122, no.~1, p.~e2408799121, 2025.

\bibitem{poggio2017and}
T.~Poggio, H.~Mhaskar, L.~Rosasco, B.~Miranda, and Q.~Liao, ``Why and when can deep-but not shallow-networks avoid the curse of dimensionality: a review,'' {\em International Journal of Automation and Computing}, vol.~14, no.~5, pp.~503--519, 2017.

\bibitem{hein2005intrinsic}
M.~Hein and J.-Y. Audibert, ``Intrinsic dimensionality estimation of submanifolds in rd,'' in {\em Proceedings of the 22nd international conference on Machine learning}, pp.~289--296, 2005.

\bibitem{spigler2020asymptotic}
S.~Spigler, M.~Geiger, and M.~Wyart, ``Asymptotic learning curves of kernel methods: empirical data versus teacher--student paradigm,'' {\em Journal of Statistical Mechanics: Theory and Experiment}, vol.~2020, no.~12, p.~124001, 2020.

\bibitem{vershynin2018high}
R.~Vershynin, {\em High-dimensional probability: An introduction with applications in data science}, vol.~47.
\newblock Cambridge university press, 2018.

\end{thebibliography}
\bibliographystyle{ieeetr}

\newpage

\newpage

\begin{appendices}
\section{Analytical model  for the optimal $\beta$ }
\label{sec:dimostrazione}

Consider the infinite data (that means that we neglect the fluctuations of the loss over variations of the training set), high (but finite) $d, n$ scenario, with labels generated according to the sign of the dot product against a teacher vector $\mathbf w_* \in \mathbb R^d$  with $|| \mathbf w_* || = 1$. Let $\mathbf W$ be a Gaussian random matrix and $\mathbf x$ a Gaussian random vector that models the data population.
The expectation (that we denote with $\bar L$), with respect to realizations of the random matrix $\mathbf W$, of the $0\text{-}1$  loss of the ensemble (that we denote with $L$) can be written as (with $\mathbbm{1}$ intended as a function that  maps a logical  statement into  its truth value, encoded  with $1$ for true and $0$ for false)
\begin{equation*}
    \begin{split}
    & \bar{L} := 
    \mathbb E_{\mathbf x, \mathbf W}
    \mathbbm{1}
    \Bigg[ \!
    \operatorname{sign} 
    \!
    \left(
    \!
    \sum_i^n 
        \exp \left( 
            - \beta 
              \mathbb E_{\mathbf x'}
              \mathbbm{1}
              \left[
              \operatorname{sign}( \mathbf w_*^T \mathbf x') 
              \neq \operatorname{sign}( \mathbf w_i^T\mathbf x'   )
              \right]
            \right)
              \operatorname{sign}(\mathbf x^T \mathbf w_i )
              \! \!
    \right)
    \! \!
    \neq
    \!
     \operatorname{sign}(\mathbf x^T \mathbf w_*)
    \Bigg]
    \end{split}
\end{equation*}
with the adoption of the expectation with respect to $\mathbf x$ and $\mathbf x'$ (which are identically distributed, as isotropic multivariate normal random vectors) in place of the sample mean with respect the observed data points thanks to the infinite $N$ assumption.
Note that also individual perceptrons $\mathbf x \mapsto \mathbf w_i^T \mathbf x $ employ the $0\text{-}1$ loss to quantify their risk measure.
We aim, in this context, to build a mathematical model  to describe the optimal value of $\beta$, such that the classification of the ensemble is optimal respect to the $0\text{-}1$ loss, with the further challenge that such loss is not differentiable by definition.
We start our argument by observing  that
\begin{equation*}
    \mathbbm{1}[x \neq y] = - \frac{1}{2}x y + \frac{1}{2}
    \hspace{2em} \forall x,y \in  \{-1,1\}
\end{equation*}
and we can thus  rewrite
\begin{equation*}
    \begin{split}
    & \bar L = 
    -
    \frac{1}{2}
    \mathbb E_{\mathbf x, \mathbf W}
    \Bigg[ 
    \operatorname{sign} \Bigg\{
    \sum_i^n 
        \exp \left( 
            - \beta 
              \mathbb E_{\mathbf x'}
              \left[
              -
              \frac{1}{2}
              \operatorname{sign}( \mathbf w_*^T \mathbf x') 
              \operatorname{sign}(  \mathbf w_i^T
              \mathbf x')
              + \frac{1}{2}
              \right] 
            \right) \times 
            \\ & \hspace{15em} \times
              \operatorname{sign}(\mathbf x^T \mathbf w_i )
    \Bigg\}
     \operatorname{sign}(\mathbf x^T \mathbf w_*)
    \Bigg]
    + \frac{1}{2}.
    \end{split}
\end{equation*}

By a substitution ($\tilde \beta = \beta/2$) and an affine transformation $x \mapsto 2 (- x + 1/2)$ we can rewrite, casting the problem from a minimization to a maximization (note that affine transformations cannot alter the number or the position of critical points), as

\begin{equation*}
    - \bar L \propto \mathbb E_{\mathbf x, \mathbf W}
    \left[
    \operatorname{sign} \left(
    \sum_i^n 
        \exp \left( 
            \tilde \beta 
              \mathbb E_{\mathbf x'}
              \left[
              \operatorname{sign}( \mathbf w_*^T \mathbf x' {\mathbf x'}^T \mathbf w_i  )
              \right]
            \right)
              \operatorname{sign}(\mathbf w_*^T \mathbf x \mathbf x^T \mathbf w_i )
    \right)
    \right],
\end{equation*}
which can be rewritten, employing the identity $\operatorname{sign}(x) = 2 \cdot  \mathbbm{1}_{\ge}(x) - 1$ that holds for every $x \neq 0$, as
\begin{equation*}
    - \bar L 
    \propto
    \mathbb E_{\mathbf x, \mathbf W}
    \mathbbm{1}
    \left[
    \frac{1}{n}
    \sum_i^n 
        \exp \left( 
             \tilde \beta 
              \mathbb E_{\mathbf x'}
              \left[
              \operatorname{sign}(  \mathbf w_*^T \mathbf x' {\mathbf x'}^T \mathbf w_i  )
              \right]
            \right)
              \operatorname{sign}(\mathbf w_*^T \mathbf x \mathbf x^T \mathbf w_i )
    \ge 0 
    \right]
\end{equation*}
and employing the equality $\mathbb E[\mathbbm{1}_{\Omega}(x)] = \mathbb P[x \in \Omega]$ we can rewrite

\begin{equation*}
    - \bar L \propto
    \mathbb P_{\mathbf x, \mathbf W}
    \left[
    \frac{1}{n}
    \sum_i^n 
        \exp \left( 
             \tilde \beta 
              \mathbb E_{\mathbf x'}
              \left[
              \operatorname{sign}( \mathbf w_*^T  \mathbf x' {\mathbf x'}^T \mathbf w_i  )
              \right]
            \right)
              \operatorname{sign}(\mathbf w_*^T \mathbf x \mathbf x^T \mathbf w_i )
    \ge 0 
    \right].
\end{equation*}
Let us, therefore, define the family of random variables $\{Y_i\}_{i=1}^N$ as
\begin{equation*}
    Y_i = \exp \left( 
             \tilde \beta 
              \mathbb E_{\mathbf x'}
              \left[
              \operatorname{sign}( \mathbf w_*^T \mathbf x' {\mathbf x'}^T \mathbf w_i  )
              \right]
            \right)
              \operatorname{sign}(\mathbf w_*^T \mathbf x \mathbf x^T \mathbf w_i ) \ \ \forall i \in [N]
\end{equation*}
and, since $\mathbf w_i \overset{iid}{\sim} \mathbf w_j \ \forall i,j$ (by construction of the random matrix $\mathbf W$), the family $\{Y_i\}_{i = 1}^N$ can be considered as a set of identically distributed random variables, copies of $Y := Y_1$.

We can therefore study the sum as a normally distributed random variable, thanks to the Central Limit Theorem that states the convergence of distribution of the sum
\begin{equation*}
   \lim_{n \rightarrow \infty} \frac{1}{n} \sum_i^n Y_i \sim \mathcal N\left( \mathbb E_{\mathbf x,  \mathbf w}  Y, \frac{1}{n} \operatorname{Var}_{\mathbf x,  \mathbf w}  Y  \right),   
\end{equation*}
which validates the normal approximation of $n^{-1} \sum_i^n Y_i$ for large numbers of classifiers.
To describe the distribution of such sum, we need to describe its two parameters. The mean, in particular, is given by
\begin{align*}
    & \mathbb E_{\mathbf x,  \mathbf w}  [ Y ] =
    \\& =
    \mathbb E_{\mathbf x,  \mathbf w}
     \exp \left( 
             \tilde \beta 
              \mathbb E_{\mathbf x'}
              \left[
              \operatorname{sign}( \mathbf w_*^T \mathbf x' {\mathbf x'}^T \mathbf w  )
              \right]
            \right)
              \operatorname{sign}(\mathbf w_*^T \mathbf x \mathbf x^T \mathbf w )
    \\
     & =
    \mathbb E_{ \mathbf w} \bigg\{
     \exp \left( 
            \tilde \beta 
              \mathbb E_{\mathbf x'}
              \left[
              \operatorname{sign}( \mathbf w_*^T \mathbf x' {\mathbf x'}^T \mathbf w  )
              \right]
            \right) \times 
            \\ & \hspace{6em} \times  
            \mathbb E_{\mathbf x}
            \left[
              \operatorname{sign}(\mathbf w_*^T \mathbf x \mathbf x^T \mathbf w )
            \right]
            \bigg\}
    \\
     & =
    \mathbb E_{ \mathbf w} \left\{
    \partial_{\tilde \beta}
     \exp \left( 
            \tilde \beta
              \mathbb E_{\mathbf x'}
              \left[
              \operatorname{sign}( \mathbf w_*^T \mathbf x' {\mathbf x'}^T \mathbf w  )
              \right]
            \right)
     \right\}
     & \partial_\alpha \exp(\alpha x) = \exp(\alpha x) x
    \\
     & =
    \partial_{\tilde \beta}
    \mathbb E_{ \mathbf w} \left\{
     \exp \left( 
            \tilde \beta
              \mathbb E_{\mathbf x'}
              \left[
              \operatorname{sign}( \mathbf w_*^T \mathbf x' {\mathbf x'}^T \mathbf w  )
              \right]
            \right)
     \right\}
     & \text{Dominated Convergence Theorem.}
    \\
    \intertext{
    We  can study the argument of  the exponential in an analytic way employing the Grothendieck's Identity, which explains the expectation of the left and right product against the rank $1$ matrix given by the outer product of a Gaussian vector against itself by means of the cosine similarity between the left and the right multiplying vector  (remember  that, by Hypothesis, $|| \mathbf w_* || = 1$)
    }
     & =
    \partial_{\tilde \beta}
    \mathbb E_{ \mathbf w} \left\{
     \exp \left( 
            \tilde \beta
              \frac{2}{\pi} \operatorname{arcsin} 
              \left(
              \frac{1}{||\mathbf w||}
              \mathbf w_*^T  \mathbf w  \right)
            \right)
     \right\}
     & \text{Grothendieck's Identity \cite{vershynin2018high}.}
    \\
    \intertext{
    We then linearize the $\arcsin$ function by noting that the variance of its argument vanishes in higher dimensions, allowing us to approximate the trigonometric function with its (linear) behavior around the origin. We then employ a concentration result relative to the inverse norm of a Gaussian vector to obtain a normally distributed argument of the exponential  ($Z$ in this context denotes $Z \sim \mathcal N(0,1)$ and should be not confused with the definition of the partition function $Z_n$ that we employed in the introduction 
 of our model). We remand to Appendix \ref{sec:linearization} for further details on the logic behind our approximation.
    }
     & \approx
    \partial_{\tilde \beta}
    \mathbb E_{ \mathbf w} \left\{
     \exp \left( 
            \tilde \beta
              \frac{2}{\pi} 
              \frac{1}{||\mathbf w||}
              \mathbf w_*^T  \mathbf w  
            \right)
     \right\}
    \\
     & \approx
    \partial_{\tilde \beta}
    \mathbb E_{ \mathbf w} \left\{
     \exp \left( 
            \tilde \beta
              \frac{2}{\pi} 
              \mathbb E \left[ \frac{1}{||\mathbf w||} \right]
              \mathbf w_*^T  \mathbf w  
            \right)
     \right\}
    \\
     & \approx
    \partial_{\tilde \beta}
    \mathbb E_{ \mathbf w} \left\{
     \exp \left( 
            \tilde \beta
              \frac{2}{\pi \sqrt{d - 2}} 
              \mathbf w_*^T  \mathbf w  
            \right)
     \right\}
     & \text{See Appendix \ref{sec:linearization}}
    \\
     & =
    \partial_{\tilde \beta}
    \mathbb E_{Z} \left\{
     \exp \left( 
            \tilde \beta
              \frac{2}{\pi \sqrt{d - 2}} 
              Z
            \right)
     \right\} & \mathbf w^T \mathbf w_* \sim \mathcal N (0, ||  \mathbf w_* ||^2), \ ||  \mathbf w_* || = 1.
    \\
    \intertext{Finally, we can employ the well known Moment Generating Function of the Normal distribution to construct an analytic expression of the mean of the distribution}
     & =
    \partial_{\tilde \beta}
     \exp \left\{ \frac{1}{2}\left( 
            \tilde \beta
              \frac{2}{\pi \sqrt{d - 2}} 
            \right)^2
    \right \}
    & \text{M.G.F of the Normal distribution}
    \\
     & =
     \exp \left\{ \frac{1}{2}\left( 
            \tilde \beta
              \frac{2}{\pi \sqrt{d - 2}} 
            \right)^2
    \right \}
    \tilde \beta
    \left( 
              \frac{2}{\pi \sqrt{d - 2}} 
    \right)^2
    \\
     &  =: \mu.
\end{align*}

The variance, instead, is obtained --- with the same theoretical tools --- as

\begin{align*}
    \operatorname{Var}_{\mathbf x,  \mathbf w}  [Y] & =
    \mathbb E_{\mathbf x,  \mathbf w}[Y^2] - \mu^2
    \\
    & =
    \mathbb E_{\mathbf x,  \mathbf w} \left\{ 
    \exp \left( 
            2 \tilde \beta 
              \mathbb E_{\mathbf x'}
              \left[
              \operatorname{sign}( \mathbf w_*^T \mathbf x' {\mathbf x'}^T \mathbf w  )
              \right]
            \right)\left(
              \operatorname{sign}(\mathbf w_*^T \mathbf x \mathbf x^T \mathbf w )\right)^2
    \right\} - \mu^2
    \\
    & =
    \mathbb E_{\mathbf w} \left\{
    \exp \left( 
            2 \tilde \beta 
              \mathbb E_{\mathbf x'}
              \left[
              \operatorname{sign}( \mathbf w_*^T \mathbf x' {\mathbf x'}^T \mathbf w  )
              \right]
            \right)
    \right\}- \mu^2
    \\
    & \approx
    \mathbb E_{\mathbf w}  \left\{ 
    \exp \left( 
            2 \tilde \beta 
              \frac{2}{\pi \sqrt{d - 2}} 
              \mathbf w_*^T \mathbf w
            \right)
    \right\} - \mu^2
    & \text{See Appendix \ref{sec:linearization}}
    \\
    & =
    \exp 
    \left(
    \frac{1}{2} \left( 
            2 \tilde \beta 
              \frac{2}{\pi \sqrt{d - 2}} 
            \right)^2
    \right)- \mu^2
    \\
    & =
    \exp 
    \left(
    \frac{1}{2} 
              \frac{16 \tilde \beta^2 }{\pi^2 (d - 2)} 
    \right)- \mu^2
    \\
    & =
    \exp 
    \left(
              \frac{8 \tilde \beta^2 }{\pi^2 (d - 2)} 
    \right) - \mu^2
    \\
    & =: \sigma^2.
\end{align*}
Thus, we can study the probability of positive value of the average among $Y_1,...,Y_n$ as
\begin{equation*}
    \begin{split}
    \mathbb P_{\mathbf Y} \left[ \frac{1}{n} \sum_i Y_i \ge 0 \right]
    & \approx
     \mathbb P_{Z} \left[ \mu + \frac{1}{\sqrt n}\sigma Z \ge 0 \right]
    \\
    & =
    \mathbb P_{Z} \left[  Z \ge - \frac{\mu}{\sigma} \sqrt{n} \right]
    \\
    & =
    \mathbb P_{Z} \left[  - Z \le \frac{\mu \sqrt{n}}{\sigma} \right] 
    \\
    & =
    \Phi \left( \frac{\mu \sqrt{n}}{\sigma} \right) ,
    \end{split}
\end{equation*}
where $\Phi$ is the cumulative density function of the normal distribution.
Thus we get
\begin{equation*}
    - \bar L \propto
    \Phi \left( 
    \sqrt{n}
    \frac{
    \exp \left\{ \frac{1}{2}\left( 
            \tilde \beta
              \frac{2}{\pi \sqrt{d - 2}} 
            \right)^2
    \right \}
    \tilde \beta
    \left( 
              \frac{2}{\pi \sqrt{d - 2}} 
    \right)^2
    }{
    \sqrt{ 
    \exp 
    \left(
              \frac{8 \tilde \beta^2 }{\pi^2 (d - 2)} 
    \right) 
    -
    \left( 
         \exp \left\{ \frac{1}{2}\left( 
            \tilde \beta
              \frac{2}{\pi \sqrt{d - 2}} 
            \right)^2
    \right \}
    \tilde \beta
    \left( 
              \frac{2}{\pi \sqrt{d - 2}} 
    \right)^2
    \right)^2
    }
    }
    \right)
\end{equation*}
and after  applying conversely $\beta = 2 \tilde \beta$

\begin{equation*}
    - \bar L \propto 
    \Phi \left( 
    \sqrt{n}
    \frac{
    \exp \left\{ \frac{1}{2}\left( 
            \frac{\beta}{2}
              \frac{2}{\pi \sqrt{d - 2}} 
            \right)^2
    \right \}
    \frac{\beta}{2}
    \left( 
              \frac{2}{\pi \sqrt{d - 2}} 
    \right)^2
    }{
    \sqrt{ 
    \exp 
    \left(
              \frac{8 \left(\frac{\beta}{2}\right)^2 }{\pi^2 (d - 2)} 
    \right) 
    -
    \left( 
         \exp \left\{ \frac{1}{2}\left( 
            \frac{\beta}{2}
              \frac{2}{\pi \sqrt{d - 2}} 
            \right)^2
    \right \}
    \frac{\beta}{2}
    \left( 
              \frac{2}{\pi \sqrt{d - 2}} 
    \right)^2
    \right)^2
    }
    }
    \right)
\end{equation*}
and since the function $\Phi$ is monotonically increasing we obtain the final result
\begin{equation}
    \label{eq:beta_opt}
    \beta^*  
    =
    \mathop{\operatorname{argmax}}_{\beta}
    \xi(\beta)
    = 
    \mathop{\operatorname{argmax}}_{\beta}
\frac{
    \exp \left\{ \frac{1}{2}\left( 
              \frac{\beta}{\pi \sqrt{d - 2}} 
            \right)^2
    \right \}
     \beta 
    \left( 
              \frac{1}{\pi \sqrt{d - 2}} 
    \right)^2
    }{
    \sqrt{ 
    \exp 
    \left(
              \frac{ 2 \beta^2  }{\pi^2 (d - 2)} 
    \right) 
    -
    4
         \exp \left\{ \left( 
              \frac{\beta}{\pi \sqrt{d - 2}} 
            \right)^2
    \right \}
    \beta^2
    \left( 
              \frac{1}{\pi \sqrt{d - 2}} 
    \right)^4
    }
    }
\end{equation}
that leads to the minimization of the expected loss $\bar L$.
%
%


%
%

\section{Analytical construction of the maximizer of $\xi$}
\label{sec:num_an}
The derivative of $\xi(\beta)$ with respect to $\beta$ is given by the expression
\begin{equation*}
\begin{split}
\partial_\beta \xi(\beta)
& = 
\frac{\left( 2 {{\ensuremath{\pi} }^{2}} d-2 {{\beta}^{2}}-4 {{\ensuremath{\pi} }^{2}}\right) }{\left( \left( {{\ensuremath{\pi} }^{4}} {{d}^{2}}-4 {{\ensuremath{\pi} }^{4}} d+4 {{\ensuremath{\pi} }^{4}}\right)  {{\exp}\left({\frac{{{\beta}^{2}}}{{{\ensuremath{\pi} }^{2}} d-2 {{\ensuremath{\pi} }^{2}}}}\right)}-4 {{\beta}^{2}}\right)} \times \\ 
& \hspace{1em} \times 
\frac{ \sqrt{{{\ensuremath{\pi} }^{4}} {{d}^{2}}-4 {{\ensuremath{\pi} }^{4}} d+4 {{\ensuremath{\pi} }^{4}}} {{\exp}\left({\frac{{{\beta}^{2}}}{2 {{\ensuremath{\pi} }^{2}} d-4 {{\ensuremath{\pi} }^{2}}}+\frac{{{\beta}^{2}}}{{{\ensuremath{\pi} }^{2}} d-2 {{\ensuremath{\pi} }^{2}}}}\right)}}{\sqrt{\left( {{\ensuremath{\pi} }^{4}} {{d}^{2}}-4 {{\ensuremath{\pi} }^{4}} d+4 {{\ensuremath{\pi} }^{4}}\right)  {{\exp}\left({\frac{2 {{\beta}^{2}}}{{{\ensuremath{\pi} }^{2}} d-2 {{\ensuremath{\pi} }^{2}}}}\right)}-4 {{\beta}^{2}} {{\exp}\left({\frac{{{\beta}^{2}}}{{{\ensuremath{\pi} }^{2}} d-2 {{\ensuremath{\pi} }^{2}}}}\right)}}},
\end{split}
\end{equation*}
which, due to the factor $2 ( \pi^2d - 2 \pi^2 - \beta^2)$, is $0$ for
\begin{equation*}
    \beta = \pm \sqrt{\pi^2(d - 2)} = \pm \pi \sqrt{d - 2}.
\end{equation*}
Since we are constrained to the case $\beta > 0$, the solution becomes 
\begin{equation*}
    \beta^* = \pi \sqrt{d - 2},
\end{equation*}
which is experimentally compatible with minima in the loss profiles for Gaussian data.

\section{Linearization of the $\arcsin$ and approximation of the uniform distribution over the sphere}
\label{sec:linearization}
\begin{figure}[htbp]
    \centering
    \begin{subfigure}[b]{0.45\textwidth}
        \centering
        \includegraphics[scale = \scalefactor]{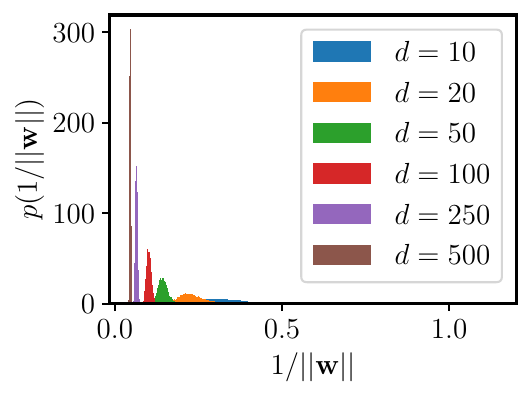}
        \caption{
 Concentration of the  distribution \\ of the image of a normal \\ random vector through the inverse norm function
 \\
        }
        \label{fig:subfig1}
    \end{subfigure}
    \hfill
    \begin{subfigure}[b]{0.45\textwidth}
        \centering
        \includegraphics[scale = \scalefactor]{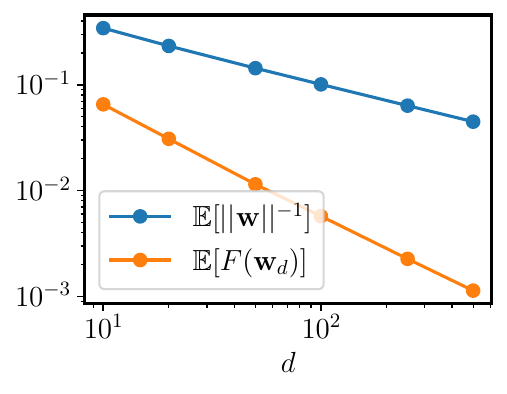}
        \caption{
         Comparison of the speed of decay 
    \\
    of the expected  inverse norm
    \( \mathbb E || \mathbf w_d ||^{-1}  \)
    \\
    and the speed of decay of the expected fluctuation
    \\
    \(
    \mathbb E 
    {
    \left| 
    {|| \mathbf w_d ||}^{-1} - 
    \mathbb E \left[ {|| \mathbf w_d ||}^{-1} \right] 
    \right |
    }
    \)
        }
        \label{fig:subfig2}
    \end{subfigure}
    
    \caption{Numerical study of the concentration properties of the inverse norm of a normal random vector}
        \label{fig:two_subplots}
\end{figure}

\begin{figure}
    \centering
    \includegraphics[scale = \scalefactor]{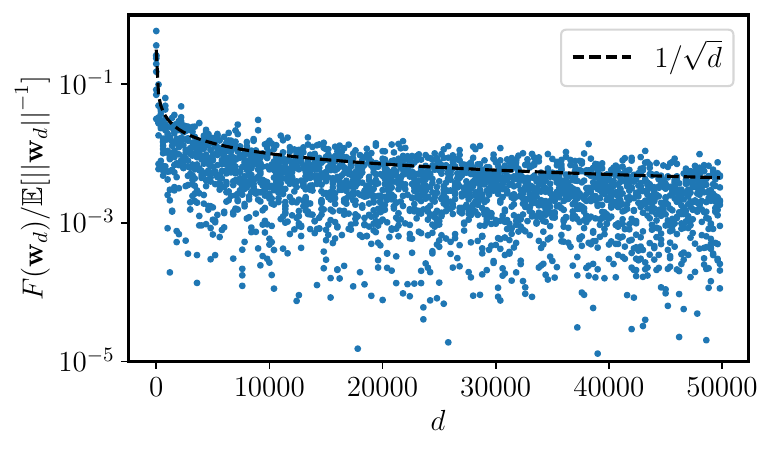}
    \caption{Observed magnitude of the normalized fluctuations of  the inverse norm. }
    \label{fig:normfluct}
\end{figure}

In Appendix \ref{sec:dimostrazione} we employed the approximation
\begin{equation*}
    \mathbb E_{ \mathbf w} \left\{
     \exp \left( 
            \tilde \beta
              \frac{2}{\pi} \operatorname{arcsin} 
              \left(
              \frac{1}{||\mathbf w||}
              \mathbf w_*^T  \mathbf w  \right)
            \right)
     \right\}
     \approx 
      \mathbb E_{ \mathbf w} \left\{
     \exp \left( 
            \tilde \beta
              \frac{2}{\pi \sqrt{d - 2}} 
              \mathbf w_*^T  \mathbf w  
            \right)
     \right\}
\end{equation*}
where the random vector $\mathbf w$ is assumed to be normally distributed with $\mathbf w  \sim \mathcal N(\mathbf 0, \mathbf I_d)$.
The first step to understand the approximation is the study  of the random variable
\begin{equation*}
     A := \frac{1}{||\mathbf w||}
              \mathbf w_*^T  \mathbf w  .
\end{equation*}
Since $\mathbf w \sim \mathcal N(\mathbf 0, \mathbf I)$ we obtain that
\begin{equation*}
    \mathbf w_*^T \mathbf w \sim \mathcal N \left( \mathbf 0,  || \mathbf w_* ||^2 \right) , 
\end{equation*}
but being the teacher vector of unitary norm with $||\mathbf w_*|| = 1$ the scalar product behaves as a normal random variable. To study the distribution of $A$ it is still necessary, however, to understand the impact of the term $\frac{1}{|| \mathbf w ||}$ to the statistics. If $\mathbf w_d \sim \mathcal N(\mathbf 0, \mathbf I_d)$ then, defining the fluctuation as
\begin{equation}
    F(\mathbf w_d) = 
    {
    \left| 
    \frac{1}{|| \mathbf w_d ||} - 
    \mathbb E \left[ \frac{1}{|| \mathbf w_d ||} \right] 
    \right |
    },
\end{equation}
we claim that
\begin{equation*}
    \lim_{d \rightarrow \infty}
    \mathbb E \left[ 
    \frac{F(\mathbf w_d)}{ \mathbb E \left[ \frac{1}{|| \mathbf w_d || } \right] }
    \right]
    =  0,
\end{equation*}
meaning that the expected fluctuation around the mean of the random variable $\frac{1}{|| \mathbf w_i ||}$ vanishes faster than its mean (as shown in Figure \ref{fig:two_subplots} and Figure \ref{fig:normfluct}), allowing us to employ, for large $d$, the approximation
\begin{equation*}
    \frac{1}{|| \mathbf w_d ||} 
    \approx
    \mathbb E \left[ \frac{1}{|| \mathbf w_d ||} \right] 
\end{equation*}
treating the inverse norm as a constant, or to be more precise as a concentrating quantity.

%
%

%
%


We can employ the approximation to compute the expectation, by considering
\begin{align*}
    \mathbb E \left[
    \frac{1}{|| \mathbf w_i ||}
    \right]
    & = 
    \mathbb E \left[
    \sqrt { 
    \frac{1}{|| \mathbf w_i ||^2}
    }
    \right]
    \\
    \intertext{and, under the hypothesis of concentration of the probability distribution, since
    $f\left( \int_{\mathbb R^d} d \mathbf x \ \delta( \mathbf x - \bar {\mathbf x} )  \mathbf x \right) = 
    \left( \int_{\mathbb R^d} d \mathbf x \ \delta( \mathbf x - \bar {\mathbf x} )  f ( \mathbf x ) \right)$, then}
    \mathbb E \left[
    \frac{1}{|| \mathbf w_i ||}
    \right] & \approx 
    \sqrt { 
    \mathbb E \left[
    \frac{1}{|| \mathbf w_i ||^2}
    \right]
    } 
    \intertext{
    and noting that $\frac{1}{|| \mathbf w_i ||^2} \sim \chi^{-2}(d)$ --- where $\chi^{-2}(d)$ denotes the inverse chi square distribution with $d$ degrees of freedom --- we obtain
    }
     \mathbb E \left[
    \frac{1}{|| \mathbf w_i ||}
    \right]
    & \approx 
    \sqrt { 
    \frac{1}{d - 2}
    }  .
\end{align*}

This implies that
\begin{equation*}
    A \sim \mathcal N \left( \mathbf 0, \frac{1}{d - 2} \right)
\end{equation*}
meaning that, for  large $d$, the fluctuations around the mean are of order $\mathcal O( \sqrt{d^{-1}})$ and allowing us to approximate the $\operatorname{arcsin}$ function with its Taylor expansion around the origin 
\begin{equation*}
    \operatorname{arcsin} \left( A \right) \approx A 
\end{equation*}
with the approximation improving as $d$ grows. This result enforces the idea that 
\begin{equation*}
    \frac{\mathbf w_d}{|| \mathbf w_d ||} \sim \mathcal{U}(\mathbb S_{d - 1}) \approx \mathcal N \left(
    \mathbf 0, \sqrt{\frac{1}{d-2}}\mathbf I_d
    \right)
\end{equation*}
that is to say, the (scaled) high dimensional normal distribution becomes --- when $d$ grows ---  a valid approximation of the uniform distribution over the unit sphere  (which is defined by means of a normalized Gaussian vector). This is a well known fact in the literature of Concentration of Measure \cite{vershynin2018high} and High Dimensional Geometry and was for us a valid tool for constructing our solution of the optimal $\beta$ problem. 

\section{Perceptron merging}
\label{sec:inference}
A similar idea to the one proposed in Appendix \ref{sec:linearization} can be employed for averaging the contribution of the perceptrons directly in the parameters space rather than in the functional space where the classifiers exist, imposing that
\begin{equation*}
    \operatorname{sign}\left( \sum_i^N 
    \alpha_i
    \phi \left( \mathbf w_i \right) \right)
    \approx 
    \phi\left( \sum_i^N \alpha_i \mathbf w_i \right)
\end{equation*}
with the appealing property that 
\begin{equation*}
     \sum_i^N \alpha_i \mathbf w_i =: \mathbf w_{ens} \in \mathbb R^d
\end{equation*}
reducing the parallel workload during the inference phase to the computation of the dot product $\mathbf x^T \mathbf w_{ens}$. This approximation can be easily obtained by noting that the dot product can be decomposed into the product of two random variables
\begin{equation*}
    \mathbf w^T \mathbf x = \operatorname{sign} \left(  \mathbf w^T \mathbf x  \right) 
    \left| \mathbf w^T \mathbf x  \right|
\end{equation*}
where we denote
\begin{equation*}
    \begin{cases}
        A(\mathbf x) \sim \left| \mathbf w^T \mathbf x  \right| \\ 
        B(\mathbf x) \sim \operatorname{sign} \left(  \mathbf w^T \mathbf x  \right)
    \end{cases}
\end{equation*}
and thus, we can imagine $A$ and $B$ as random functions from $\mathbb R^d \rightarrow \mathbb R$. Conceptually, the idea is to treat the transpose vector $\mathbf w^T$  as a random linear and continuous functional from the dual space of $\mathbb R^d$ to $\mathbb R$ and to decompose it into the product of two random non linear functionals ($A$ and $B$).
Let us assume, for the sake of the approximation, that the number of classifiers is very high, and we can approximate the sample average of classifiers with their expected value. Note that, in this case, $\beta$ is fixed and thus we can --- \textit{a posteriori} --- neglect the fluctuations (which are, in our theory, responsible of the determination of $\beta$). 
Hence we can rewrite
\begin{equation*}
    \begin{split}
       y^{lin}_{ens}(\mathbf x) & :=  \operatorname{sign}
        \left( 
        \frac{1}{n} 
        \sum_i^n
        \left[
        \exp \left( 
             \tilde \beta 
              \mathbb E_{\mathbf x'}
              \left[
              \operatorname{sign}( \mathbf w_*^T  \mathbf x' {\mathbf x'}^T \mathbf w_i  )
              \right]
            \right)
              \mathbf x^T \mathbf w_i
        \right]
        \right)
        \\
        & \approx
        \operatorname{sign}
        \left( 
        \mathbb E_{\mathbf w}
        \left[
        \exp \left( 
             \tilde \beta 
              \mathbb E_{\mathbf x'}
              \left[
              \operatorname{sign}( \mathbf w_*^T  \mathbf x' {\mathbf x'}^T \mathbf w  )
              \right]
            \right)
              \mathbf x^T \mathbf w 
        \right]
        \right)
        \\
        & =
        \operatorname{sign}
        \left( 
        \mathbb E_{A,B}
        \left[
        \exp \left( 
             \tilde \beta 
              \mathbb E_{\mathbf x'}
              \left[
              \operatorname{sign}( \mathbf w_*^T  \mathbf x' B(\mathbf x')  )
              \right]
            \right)
              B(\mathbf x)
              A(\mathbf x)
        \right]
        \right).
        \\
\end{split}
\end{equation*}
Note that the term $A(\mathbf x')$ in the exponential has been neglected since it is composed with the $\operatorname{sign}$ function. Under the assumption of statistical independence between $A$ and $B$ we can rewrite
\begin{equation*}
\begin{split}
        y^{lin}_{ens} & \approx
        \operatorname{sign} \left( 
        \mathbb E_{B}
        \left[
        \exp \left( 
             \tilde \beta 
              \mathbb E_{\mathbf x'}
              \left[
              \operatorname{sign}( \mathbf w_*^T  \mathbf x' B(\mathbf x') )
              \right]
            \right)
              B(\mathbf x)
        \right]
        \mathbb E_{A}
        \left[
              A(\mathbf x)
        \right]
        \right)
    \end{split}
\end{equation*}
and since $\mathbb E_A[A(\mathbf x)] = \mathbb E_{\mathbf w} |\mathbf w^T  \mathbf x | $ is a positive constant we can rewrite
\begin{equation*}
    \begin{split}
        y^{lin}_{ens} & \approx \operatorname{sign} \left( 
        \mathbb E_{B}
        \left[
        \exp \left( 
             \tilde \beta 
              \mathbb E_{\mathbf x'}
              \left[
              \operatorname{sign}( \mathbf w_*^T  \mathbf x' B(\mathbf x') )
              \right]
            \right)
              B(\mathbf x)
        \right]
        \right)
        \\
        & \approx 
        \operatorname{sign}
        \left( 
        \frac{1}{n} 
        \sum_i^n
        \left[
        \exp \left( 
             \tilde \beta 
              \mathbb E_{\mathbf x'}
              \left[
              \operatorname{sign}( \mathbf w_*^T  \mathbf x' {\mathbf x'}^T \mathbf w_i  )
              \right]
            \right)
              \operatorname{sign}\left( \mathbf x^T \mathbf w_i \right)
        \right]
        \right)
    \end{split}
\end{equation*}

for any choice of $\mathbf x$ we conclude that
\begin{equation*}
    y_{ens}(\mathbf x) \approx  y^{lin}_{ens}(\mathbf x),
\end{equation*}
with the approximation improving when $n$ is large (i.e. when the fluctuations of the sample average of classifiers become negligible). The possibility of averaging the contribution of classifiers directly in the parameter space appears as a counterintuitive result due to the strong non linearity of the realization map $\phi$, since
\begin{equation}
    \operatorname{sign}\left\{ (\alpha \mathbf w_1^T +  \beta \mathbf w_2^T)    \mathbf x\right\}
    \neq 
    \alpha \operatorname{sign}\left(  \mathbf w_1^T \mathbf x\right) + \beta \operatorname{sign}\left(  \mathbf w_2^T \mathbf x\right),
\end{equation}
with the latter term possibly being even outside of the codomain of the $\operatorname{sign}$ function. This property offers a significant computational advantage (since we can approximately merge the ensemble of perceptrons into a single perceptron, consuming fewer resources) and is, supposedly, a remarkable consequence of the symmetries in the geometry of the Isotropic Gaussian distribution. A minimal working example in Python is proposed in Appendix \ref{sec:minwork}.

\section{Minimal Working Example}
\label{sec:minwork}
We provide a minimal Python code that may be used as a model for the reproduction of the experiments listed across our work. The code is shown in Code Snippet \ref{lst:mwe}.

\renewcommand{\lstlistingname}{Code Snippet}
\begin{lstlisting}[style=pythonstyle, label = lst:mwe, caption = Minimal Working Example]
import jax
import numpy as np
from jax.numpy import sign, exp, ones
from jax.numpy.linalg import norm
from jax.random import PRNGKey, normal

# Hyperparameters
N          = 10000           # number of samples
d          = 500             # dimensions
n          = 20000           # number of classifiers
train_seed = 42              # simulation seed
test_seed  = 777                
w_star     = ones(d)         # replace with any teacher vector
w_star    /= norm(w_star)    # the teacher has unitary norm

# Generation of data (Train)
X = normal(PRNGKey(train_seed),shape = (N,d))
Y = sign(X @ w_star)

# Generation of data (Test)
X_test = normal(PRNGKey(test_seed),shape = (N,d))
Y_test = sign(X_test @ w_star)

# Construction of random classifiers
perceptron_seed = 2025
W = normal(PRNGKey(perceptron_seed),shape = (n,d))

# compute the 0-1 loss for each classifier
loss_01  = jax.vmap(lambda w: (sign(X@w) != Y).mean() )(W)

# compute the score
beta     = np.pi * np.sqrt(d - 2)
score    = exp( - beta * loss_01 )
alfa     = score / score.sum()

# predict train
Y_hat_train = sign(sign(X@W.T) @ alfa)
Y_hat_test  = sign(sign(X_test@W.T) @ alfa)

# output
train_acc   = (Y_hat_train == Y).mean()
test_acc    = (Y_hat_test  == Y_test).mean()
print("Train accuracy : %.3f" % (train_acc) )
print("Test  accuracy : %.3f" % (test_acc) )
\end{lstlisting}

The construction of $\widehat {\mathbf y}$ (denoted in Code Snippet \ref{lst:mwe} as \texttt{Y\_hat\_train}  for train data and \texttt{Y\_hat\_test} for test data) is based on
\begin{align*}
    \widehat {\mathbf y}(\mathbf X, \mathbf W) & = \sum_i \mathbf e_i
    \operatorname{sign} \left(
    \sum_j
        \operatorname{sign} \left(
            \mathbf w_j^T \mathbf x_i
        \right)
        \alpha_j
    \right)
    \\
    & = \sum_i 
    \mathbf e_i
    \operatorname{sign} \left(
    \sum_j
        \operatorname{sign} \left(
                \mathbf e_i^T
                \mathbf X
            \mathbf w_j
        \right)
        \alpha_j
    \right)
    \\
    & = \sum_i 
    \mathbf e_i
    \operatorname{sign} \left(
    \mathbf e_i^T
    \sum_j
        \operatorname{sign} \left(
                \mathbf X
            \mathbf w_j
        \right)
        \alpha_j
    \right)
    \\
    & = \sum_i 
    \mathbf e_i
    \mathbf e_i^T
    \operatorname{sign} \left(
        \operatorname{sign} \left(
                \mathbf X
            \mathbf W^T
        \right)
       \boldsymbol{\alpha}
    \right)
    \\
    & = 
    \operatorname{sign} \left(
        \operatorname{sign} \left(
                \mathbf X
            \mathbf W^T
        \right)
       \boldsymbol{\alpha}
    \right)
\end{align*}

where the $\operatorname{sign}$ function is intended in a generalized  sense as a map from tensors (and thus also scalars) to tensors of the same dimensions whose entries are the sign of the corresponding entries in the input tensor. The resulting output is shown in Terminal \ref{lst:out}.

\renewcommand{\lstlistingname}{Terminal}
\begin{lstlisting}[style=pythonstyle,caption = Output of the Code Snippet \ref{lst:mwe}, label = lst:out]
Train accuracy : 0.901
Test  accuracy : 0.862
\end{lstlisting}

It is also possible to test the proposal of Appendix \ref{sec:inference} with few lines of code, as shown in Code Snippet \ref{lst:mwe_merge}, obtaining a result (shown in Terminal
\ref{lst:output_merge}) which is compatible with the metrics obtained for the unmerged ensemble (shown in Terminal \ref{lst:out})
\renewcommand{\lstlistingname}{Code Snippet}
\begin{lstlisting}[style=pythonstyle,caption = Output of the snippet, label = lst:mwe_merge]
w_ens = score @ W
Y_hat_train_merge = sign(X @ w_ens)
Y_hat_test_merge  = sign(X_test @ w_ens)

train_acc_merge   = (Y_hat_train_merge == Y).mean()
test_acc_merge    = (Y_hat_test_merge  == Y_test).mean()
print("Train accuracy merged perceptrons : %.3f" % (train_acc_merge) )
print("Test  accuracy merged perceptrons : %.3f" % (test_acc_merge) )
\end{lstlisting}

\renewcommand{\lstlistingname}{Terminal}

\begin{lstlisting}[style=pythonstyle,caption = Output of the snippet \ref{lst:mwe_merge}, label = lst:output_merge]
Train accuracy merged perceptrons : 0.900
Test  accuracy merged perceptrons : 0.874
\end{lstlisting}

\end{appendices}

\end{document}